\documentclass[12pt]{article}
\usepackage[letterpaper, portrait, margin=1in]{geometry}
\usepackage[utf8]{inputenc}
\usepackage{cite}
\usepackage{hyperref}

\hypersetup{
    colorlinks=true,
    linkcolor=blue,
    filecolor=magenta,      
    urlcolor=cyan,
    citecolor=black
    }
\usepackage{amsmath,amsfonts,amssymb,bm,xfrac}

\usepackage{float,caption}
\usepackage[labelformat=simple]{subcaption}
\graphicspath{{./img/}}

\usepackage{todonotes}
\usepackage[nolist]{acronym}

\usepackage{setspace}
\usepackage{parskip}
\title{NFL Career Success as Predicted by NFL Scouting Combine}

\author{Brian Szekely\textsuperscript{1}, Christian Sinnott\textsuperscript{1}, Savannah Halow\textsuperscript{1}, Gregory Ryan\textsuperscript{2}}

\date{}
\begin{document}

\begin{acronym}
\acro{NFL}{National Football League}
\acro{RMSE}{Root Mean Squared Error}
\end{acronym}

\maketitle
\begin{center}
\textbf{1} Department of Psychology, University of Nevada, Reno, Reno
\newline
\textbf{2} 
College of Nursing \& Health Sciences, Piedmont University, Demorest, Georgia
\end{center}




\section{ABSTRACT}
The \ac{NFL} Scouting Combine serves as a tool to evaluate the skills of prospective players and assess their readiness to play in the \ac{NFL}. The development of machine learning brings new opportunities in assessing the utility of the Scouting Combine. Using machine and statistical learning, it may be possible to predict future success of prospective athletes, as well as predict which Scouting Combine tests are the most important. Results from statistical learning research have been contradicting whether the Scouting combine is a useful metric for player success. In this study, we investigate if machine learning can be used to determine matriculation and future success in the \ac{NFL}. Using Scouting Combine data, we evaluate six different algorithms' ability to predict whether a potential draft pick will play a single \ac{NFL} snap (matriculation). If a player is drafted, we predict how many snaps they go on to play (success). We are able to predict matriculation with 83\% accuracy; however, we are unable to predict later success. Our best performing algorithm returns large error and low explained variance (RMSE=1,210 snaps; ${R}^2$=0.17). These findings indicate that while the Scouting Combine can predict \ac{NFL} matriculation, it may not be a reliable predictor of long-term player success.

\section{INTRODUCTION}

The \ac{NFL} is a highly competitive American football league which generates billions of dollars in revenue annually \cite{statista}. As the \ac{NFL} is the highest level of competition for American football, thousands of prospective players hope to enter its ranks every year via free agency or the \ac{NFL} Draft. Roster spots for each of the 32 \ac{NFL} teams are limited, however; teams begin with 90 players during the off-season, then reduce rosters to 53 players during the regular season. Given the extreme physical demand of the sport, the revenue each team is expected to generate, and the small number of players a team can employ; teams face high pressure to select the most skilled and physically talented players. 

To aid this, the \ac{NFL} created the Scouting Combine in 1985: a week-long event where draft prospects showcase their abilities to \ac{NFL} team management through completion of a series of physical drills. While it may seem obvious that potential \ac{NFL} players need to be strong, fast, and otherwise physically capable of playing in the \ac{NFL} \cite{Garstecki2004}, the Scouting Combine was the first league-wide attempt at evaluating potential players in this way. While useful, the Scouting Combine is not perfect. Drills performed by players in the Scouting Combine do not always correspond with success in the \ac{NFL} proper. In some cases, drills performed in the Scouting Combine can have poor correspondence with on-field player outcomes \cite{McGee2003}. 

Previous research has attempted to rigorously model the relationship between Scouting Combine measures and career outcomes including career salary, draft location, and games played in a position-wise fashion \cite{Kuzmits2008}. Sparse research exists which models Scouting Combine performance and career outcome using machine learning methods, despite their growing utility in sports research. Furthermore, little research exists investigating the relationships between Scouting Combine metrics and career outcomes across all position groups. Using five years of \ac{NFL} Scouting Combine, draft data, and career performance metrics, this work aims to determine to what extent Scouting Combine performance can predict future career performance across all positions. 

More specifically, we first aim to determine if Scouting Combine measures can predict whether a draftee will go on to play in the \ac{NFL} at all. Around 260 players are drafted each year in the \ac{NFL}, but drafting alone does not guarantee the player makes a roster. Many players who are drafted are later cut. Second, we seek to determine if Scouting Combine metrics can predict how many total snaps a player will play once they are in the \ac{NFL}. It is unclear whether performance in the Scouting Combine matters once a player matriculates into the \ac{NFL}. 

\subsection{Related Work}

Few studies have used similar methods to try and predict \ac{NFL} player outcomes, despite high interest. This may be due to both the relatively recent use of machine learning methods in sports science broadly, as well as "desirable" characteristics of athletes changing over time. Through our work we hope to add to extant literature studying player evaluation and forecasting, and establish clearer guidelines for predicting later player success. At time of writing our work will be the first to try and establish relationships between \ac{NFL} Scouting Combine performance, matriculation into the \ac{NFL}, and the number of plays an \ac{NFL} player later participates in.

One study investigating correspondence between Scouting Combine performance and draft status was conducted by McGee and Burkett in 2003. In analyses conducted by player position, researchers observed relationships between Scouting Combine performance and draft status in three of seven positions evaluated \cite{McGee2003}. In contrast, work conducted by Robbins and colleagues in 2010 showed little correspondence between Scouting Combine performance and draft position \cite{Robbins2010}. Similar work conducted in 2013 by King showed inverse trends between draft position and \ac{NFL} career performance: most wide receivers and running backs drafted between the 3rd and 7th rounds had greater average career statistics than their 1st and 2nd round contemporaries. Notably, relatively recent work conducted by Teramoto and colleagues in 2016 showed correspondence between performance in two Scouting Combine exercises (vertical jump, 40-yard dash) and in-game performance of \ac{NFL} running backs and wide receivers \cite{Teramoto2016}. Taken together, these works suggest that Scouting Combine performance may predict draft position for some, but not all players; furthermore, while Scouting Combine metrics may drive draft position, they may not predict future success.

A thesis by Meil in 2018 attempted to establish relationships between Scouting Combine metrics, post-season performance, and years played in the \ac{NFL} by \ac{NFL} position\cite{Meil20018}. Meil found that half of \ac{NFL} player positions evaluated in this way had at least one Scouting Combine metric that significantly predicted \ac{NFL} longevity and post-season performance. While this work is not conclusive on its own, especially when considered in the context of other related work, Meil used multiple regression models to evaluate whether Scouting Combine metrics could predict future performance. In order to perform a more thorough evaluation we compare models constructed using five different statistical and machine learning algorithms. Through this approach, we may better explain correspondence between Scouting Combine metrics and future \ac{NFL} player performance, and glean additional insight into the applicability of these algorithms in sports science. 

\section{METHODS}

\subsection{Pre-Processing}

Data used in this study were obtained from \ac{NFL} draft classes from the period of 2013-2017. We remove all samples with missing features, including players with incomplete or absent Scouting Combine data. Of the original 1,973 samples, 805 were used for further analysis. To evaluate performance of candidate models, we partition data into training and test sets using an 80/20 split.

\subsection{Model Selection}
We use five common classification models to predict matriculation into the NFL: support vector machine, multivariate logistic regression, gradient boosting, random forest, and decision tree. We select these models as all have been used previously (with varying success) to classify or predict sports performance in the context of game outcomes \cite{Gomez2019,Chen2021,Apostolou2019} or player performance \cite{Sudhandradevi, Mulholland2014, Teramoto2016, Apostolou2019}. We use the same (support vector machine, gradient boosting, random forest, decision tree) or similar (multivariate linear regression) regression models in order to predict future NFL snaps. These models are evaluated based on accuracy for the classification models and \ac{RMSE} for the regression models; the model which performs best, e.g. has the greatest accuracy or the lowest \ac{RMSE}, is used in our final analysis.

Our feature vector consists of measurements taken from athletes performing six exercises at the NFL Scouting Combine. These include the 40-yard dash, broad jump, bench press, vertical jump, shuttle run, and three cone drill. The 40-yard dash measures how quickly a player can sprint 40 yards, measured to the hundredth of a second. The broad jump measures the horizontal distance a player can leap from a standing position, while the vertical jump measures the vertical distance a player can jump from a standing position: both measure this distance in inches. Bench press performance is operationalized by counting the number of repetitions a player can complete when trying to bench press 225 pounds. The shuttle run and three cone drills are a set of agility drills where a player must run a pre-determined route. Performance for this drill is measured to the hundredth of a second.  

Our label is total snaps played in the \ac{NFL}. This counts the total number of plays a player in the \ac{NFL} participated in during the season. This does not distinguish during what phase of the game the player is playing (e.g. on the offensive, defensive, or special teams portion of a game). Many players only play during one phase of the game, but if they do play multiple phases, it is almost always either offense and special teams or defense and special teams.

\subsubsection{Decision Tree}

The decision tree classification model is a supervised learning approach that uses features to predict membership of a sample with a label through a series of binary decisions. These binary decisions are nodes that are bifurcated based on the ability for a feature to split the data. Decision tree regression works in a similar manner, but using a continuous, real value as a label (number of \ac{NFL} snaps taken) rather than a binary label (whether or not a player took any \ac{NFL} snaps). We chose to use this approach due to its simplicity, ease of human interpretation, and previous usage in athletics research \cite{Gomez2019}. This is particularly important because these results need to be interpretable by those in the sports medicine field who are not as familiar with these algorithms. Having an algorithm that can successfully predict performance and can be easily understood is both novel and important to the field.

\subsubsection{Gradient Boosting}
Gradient boosted classification, like the previous model, is a supervised machine learning method which ultimately provides a decision tree model. Unlike the previous model, gradient boosting uses multiple decision trees to construct a final, ensemble model. Information from previous decision trees inform construction of subsequent decision trees that minimize prediction error through algorithms such as AdaBoost which up-weight strong learners and down-weight weak learners \cite{GradientBoosting}. Gradient boosted regression uses a similar process to predict continuous, real-valued labels. Previous sports analytics research has used this to predict final scores of professional basketball games \cite{Chen2021}. By using this classifier in this experiment, we provide additional insight into the use of these algorithms in sports medicine in general.   

\subsubsection{Random Forest}
Random forest classification, like gradient boosted classification, results in an ensemble decision tree model. In contrast to gradient boosting classification that uses information from previous decision tree models to optimize future decision tree models, random forest classification constructs all decision trees at once. These decision trees are constructed with bootstrapped training data, leading to slight differences between any set of individual decision trees. Other random forest methods may randomize the features by which a tree is constructed as well. Previous work has used random forest models to predict NFL game outcomes, tennis game outcomes \cite{yousefi2019}.

\subsubsection{Logistic and Linear Regression}
Logistic and linear regression are two basic approaches that can be used to classify or predict a binary or continuous label, respectively. Each of these models attempt to fit a function to a set of data: either a logistic function which predicts membership in a binary class, or a linear function which predicts the value of a continuous label. Both of these models assume that the data are linear in nature.  Both logistic and linear regression have been used in sports analytics, including previous attempts to predict NFL success \cite{Sudhandradevi, Mulholland2014, Teramoto2016}. As these prior attempts were largely unsuccessful, revisiting this topic with new features, metrics, and datasets demonstrates to the field a better methodology of using logistic and linear regression for sports prediction.

\subsubsection{Support Vector Machine}
Support vector machine classifier is a supervised learning model that works as a non-probabilistic binary linear classifier. This algorithm maps training samples in a space to maximize the distance between two features. Similar to Logistic Regression, support vector machine algorithms have been used to predict soccer game outcomes, with relatively low error \cite{Apostolou2019}. The support vector machine's goal is to find the distance between two linearly separable points from multiple features that return the least amount of error, while maintaining perpendicular distance between the nearest points. This method of analysis in the context of sports analytics has demonstrated to have one of the highest amounts of error when predicting outcomes measures in soccer players \cite{Apostolou2019}; However, this method has not been assessed in the context of football, and has had great success at being able to categorize events from sports video data \cite{Xu2008}. Given that this algorithm is more popular in the field, applying it in this new context may demonstrate to the field extended use cases, providing additional insight into its capabilities and potential usages.

\subsection{Cross-Validation and Model Tuning}

We perform a 10-fold cross validation on all models determine which of the five classification models has the best performance on the selected features and labels. We confirm from this validation that the random forest classifier returned the lowest error and bias of our classifiers, and that similarly, linear regression also returned the lowest bias and error of our regressions (Table ~\ref{fig:Figure1}). The data and code are publicly available on GitHub (\url{https://github.com/bszek213/nfl_combine/tree/publish})
 
 \begin{table}[h]
 \begin{center}
 \caption{ Comparison of models on training data 10-fold cross validation.}\label{tab:2}
\begin{tabular}{c|c|c} 
  \textbf{Model} & \textbf{Classification} & \textbf{Regression} \\ 
    \hline
    Decision Tree  & 0.73 & 1731.3 \\ 
    \hline
    Gradient Booster &  0.77 & 1289.3 \\
    \hline
    Support Vector Machine &  0.76 & 1298.8 \\ 
    \hline
    Random Forest & 0.81 & 1276 \\
    \hline
    Logistic Regression &  0.75 & - \\
    \hline
    Linear Regression &  - & 1210.1 \\
\end{tabular}
\end{center}
\label{table1}
 \end{table}
\section{RESULTS}
The cross-validation on the training data demonstrates that the random forest classifier has the best accuracy for NFL matriculation prediction; this model has an accuracy of 0.81, with the other models performing at 0.75, 0.77, 0.76, and 0.73 for multivariate logistic regression, gradient boosting, support vector machine, and decision tree, respectively (Table \ref{tab:2}). For the prediction of NFL snaps linear regression performs the best with an \ac{RMSE} of 1,210.1, with the other models performing at 1,298.87, 1,289.3, 1,276.0, and 1731.3 for support vector machine, gradient boosting, random forest, and decision tree, respectively (Table \ref{tab:2}). Following this confirmation, we tune these models by evaluating the hyperparameters of the random forest classifier. For the random forest model, the number of estimators, which is also known as the number of trees that were created, is 944. The number of samples that required to split each tree node is set to 2, while the number of samples that required to be in each leaf is set to two (binary tree). The number of max features was the square root of the total number of input features, which was 3. The max depth of the tree was 97, with sample with replacement bootstrapping occurring during the construction of each tree. Preliminary testing demonstrates that additional iterations did not help accuracy, but cost additional computational power. 

\begin{figure}[h]
  \centering
  \includegraphics[width=1\linewidth]{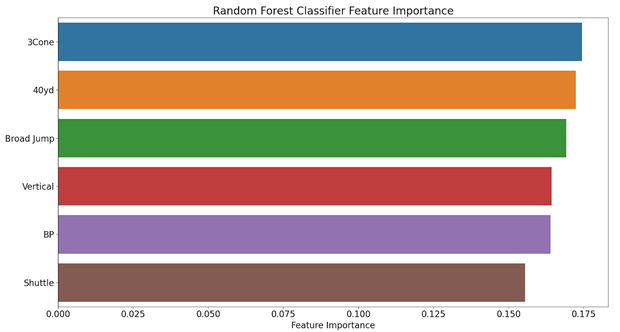}
  \caption{Feature importance as explained by the support vector machine's coefficients.}
  \label{fig:Figure1}
\end{figure}

\begin{figure}[h]
  \centering
  \includegraphics[width=1.0\linewidth]{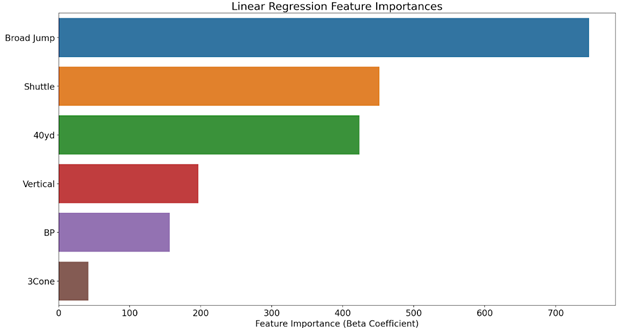}
  \caption{Feature importance as explained by the linear regression's beta coefficients.}
  \label{fig:Figure2}
\end{figure}

After tuning our models, we evaluate our model performance on the test data. We observe that the random forest classifier had a 0.83 accuracy, while the multivariate linear regression has an \ac{RMSE} of 904.6 and an ${R}^2$ of 0.17.After analyzing the explained variance represented by the feature importances of the Random Forest model and the beta coefficients obtained from the Linear Regressor, we found that the 3-cone drill and broad jump exhibit the highest importance, respectively (see Figures \ref{fig:Figure1} and \ref{fig:Figure2}).

\section{DISCUSSION}
The goal of this study was to determine if machine learning techniques can be used to effectively predict \ac{NFL} matriculation and future \ac{NFL} success using metrics from the \ac{NFL} Scouting Combine. Results from tested classification models demonstrate that we can effectively predict matriculation using a random forest classifier with an accuracy of 83\%. This result is rather intuitive, as team management makes their initial draft decision based on the Combine performance. However, it is interesting that this performance appears to correspond with the player’s likelihood of not only being drafted, but also continuing on to play in the \ac{NFL}. In terms of acceptance into the \ac{NFL}, Scouting Combine performance appears to be an effective predictor.


The same cannot be said for future success of the \ac{NFL} athlete overall. Of our regression algorithms, linear regression is the most effective for predicting \ac{NFL} snaps from the input \ac{NFL} Scouting Combine metrics. However, given the large \ac{RMSE} value of 1,210.1, we conclude that we were unable to predict future NFL career success as quantified by snaps with the given features from the Combine. This is similar to previous literature which has demonstrated that the Combine is ineffective for predicting other metrics of career success \cite{Mulholland2014,Sudhandradevi,Teramoto2016}. Taken together, this growing body of literature suggests a bleak outlook for the Combine; while performing well in the drills from the Combine may get a player into the \ac{NFL}, their performance may have little to do with how successful their \ac{NFL} career will be. If \ac{NFL} team management wants to find the best prospective athletes for their team, using additional metrics or developing new drills that correspond better with prolonged \ac{NFL} success could prove more effective.

Analysis of the feature importance of the random forest classifier and linear regression models yielded a discrepancy in importance between Scouting Combine metrics. Specifically, the 3-cone drill (Figure \ref{fig:Figure1}) carried the most weight in predicting matriculation, while the broad jump (Figure \ref{fig:Figure2}) carried the greatest weight in predicting \ac{NFL} success. This would suggest that the 3-cone drill is the most important metric or the metric perceived as most important for being drafted into the \ac{NFL}. Performance in the broad jump, however, seems to best determine how much an athlete will go on to play. In fact, the 3-cone drill may be the worst predictor of future success, as we found it had the least feature importance in our regression model (Figure \ref{fig:Figure2}). These findings are limited, however, by the overall low performance of the model. Nonetheless, our analyses suggest that \ac{NFL} scouts and coaches may be over-weighting incorrect metrics when picking athletes for their teams. Future work should investigate best metrics for evaluating athletic potential, both in the context of \ac{NFL} performance specifically as well as more generally.

Using \ac{NFL} snaps as a proxy for \ac{NFL} success comes with limitations; while we cannot predict how often an athlete will play using the Combine alone, it’s possible that performance could be an effective predictor of other “success” metrics such as yards, touchdowns, etc. which may be more relevant to evaluate performance in a position-wise context. We chose number of \ac{NFL} snaps played specifically for its broad applicability to all players regardless of position, which to the best of our knowledge, had not been investigated before. Future research may explore this topic using other metrics of success. Furthermore, the addition of other features such as anthropometric measurements (e.g. height, limb length) or body composition (e.g. percentage of lean body mass), could yield higher accuracy and a greater ability to predict \ac{NFL} success.

\section{PRACTICAL APPLICATIONS}

The results of this study are of interest to \ac{NFL} management and athletes alike. Our code, which is available for download from our GitHub repository, can serve as a guide for athletes interested in participating in the Scouting Combine. It is possible to use this algorithm to predict their own likelihood of matriculation using the classification model with 84\% accuracy. Additionally, the results of the feature importance analysis demonstrated that prospective athletes should be focusing most strongly on improving their performance in the 3-cone drill and 40-yard dash. The results of our analysis of the connection between the Combine and later snaps suggests that the Combine is not an effective metric for measuring an athlete's abilities. However, using this same algorithm, it could be possible to evaluate new or future methods which better showcase ability.


\newpage

 
 \end{document}